\documentclass[11pt]{article}

\usepackage{acl}

\usepackage{times}
\usepackage{latexsym}

\usepackage[T1]{fontenc}
\usepackage[utf8]{inputenc}

\usepackage[utf8]{inputenc}
\usepackage{geometry}
\usepackage[table]{xcolor} % 用于表格颜色，必须加 [table] 选项
\usepackage{booktabs}      % 用于高质量的表格线
\usepackage{multirow}      % 用于合并行
\usepackage{tabularx}      % 核心：用于创建自动宽度的表格，解决背景色空隙问题

\newcolumntype{Y}{>{\centering\arraybackslash}X}

\definecolor{lightgray}{HTML}{E6E6E6} % 表头浅灰色，可根据需要微调
\definecolor{lightblue}{HTML}{D7F6FF} % 目标行浅蓝色
\usepackage{microtype}

\usepackage{inconsolata}

\usepackage{graphicx}

\title{Cumulative Path-Level Semantic Reasoning for \\ Inductive Knowledge Graph Completion}

\author{
    Jiapu Wang\textsuperscript{1}, 
    Xinghe Cheng\textsuperscript{2}, 
    Zezheng Wu\textsuperscript{3}, 
    Ruiqi Ma\textsuperscript{3}, 
    Rui Wang\textsuperscript{2}, \\
    \textbf{Zhichao Yan}\textsuperscript{4}, 
    \textbf{Haoran Luo}\textsuperscript{5}, 
    \textbf{Yuhao Jiang}\textsuperscript{3}, 
    \textbf{Kai Sun}\textsuperscript{6} \\
    \textsuperscript{1}Nanjing University of Science and Technology \\
    \textsuperscript{2}Jinan University \quad
    \textsuperscript{3}Guilin University of Electronic Technology \\
    \textsuperscript{4}Shanxi University \quad
    \textsuperscript{5}Nanyang Technological University \quad
    \textsuperscript{6}Beijing University Of Technology
}

\usepackage{amsmath}    % 修复：Environment aligned undefined 和 split undefined
\usepackage{booktabs}   % 修复：Is \usepackage{booktabs} missing?
\usepackage{multirow}   % 修复：Is \usepackage{multirow} missing?
\usepackage{amsfonts}   % (建议加上) 处理数学符号
\usepackage{dsfont}
\usepackage{enumitem}
\usepackage{hhline}
\newcolumntype{Y}{>{\centering\arraybackslash}X}
\ifdefined\LaTeXML
  \renewcommand{\rowcolor}[2][]{}
\fi
\begin{document}
\maketitle
\begin{abstract}
Conventional Knowledge Graph Completion (KGC) methods aim to infer missing information in incomplete Knowledge Graphs (KGs) by leveraging existing information, which struggle to perform effectively in scenarios involving emerging entities.
Inductive KGC methods can handle the emerging entities and relations in KGs, offering greater dynamic adaptability. While existing inductive KGC methods have achieved some success, they also face challenges, such as susceptibility to noisy structural information during reasoning and difficulty in capturing long-range dependencies in reasoning paths. To address these challenges, this paper proposes the \textbf{C}umulative \textbf{P}ath-Level \textbf{S}emantic \textbf{R}easoning for inductive knowledge graph completion (CPSR) framework, which simultaneously captures both the structural and semantic information of KGs to enhance the inductive KGC task. Specifically, the proposed CPSR employs a query-dependent masking module to adaptively mask noisy structural information while retaining important information closely related to the targets. Additionally, CPSR introduces a global semantic scoring module that evaluates both the individual contributions and the collective impact of nodes along the reasoning path within KGs. %This module can accurately identify critical semantic paths for effective reasoning, capturing long-range semantic dependencies in the reasoning path. 
The experimental results demonstrate that CPSR achieves state-of-the-art performance.
\end{abstract}

\section{Introduction}
Knowledge Graphs (KGs) are pivotal in recommendation systems \cite{WeiHX2022,yan2025atomic} and natural language processing \cite{WanWQ2023}, yet their inherent incompleteness necessitates Knowledge Graph Completion (KGC) \cite{WanWG2023} to infer missing facts. 
While traditional KGC methods \cite{WanWGL2023,HoSG2018,WanWG2022} have progressed, they struggle with emerging entities not present during training. 
To address this, inductive KGC methods \cite{TerDH2020,CheHW2021,ZhuZX2021} have been developed to handle unseen entities by learning generalizable representations from graph structures.

\begin{figure}[t]
\centering
\includegraphics[scale=0.55]{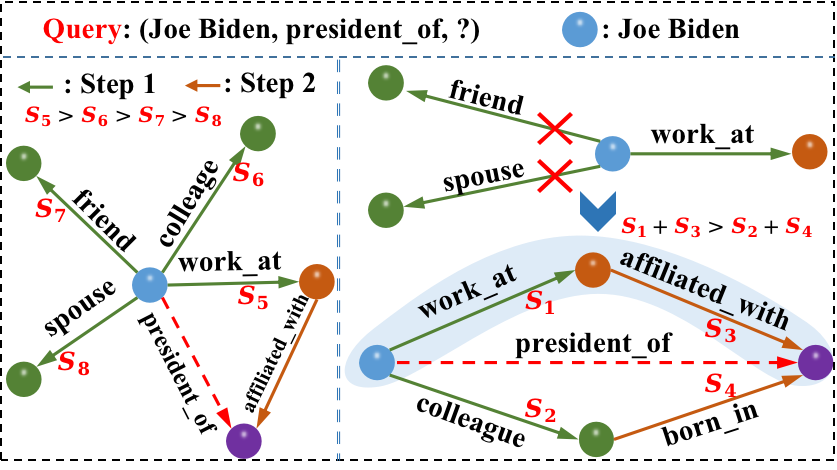}
\caption{The motivation of the conventional inductive KGC (left) and the proposed CPSR (right). 
Conventional inductive KGCs traverse nodes directly over KGs. 
CPSR sequentially masks the noise structure and captures the long-range semantic dependencies over the whole reasoning path.}
\label{path}
\end{figure}

Inductive KGC primarily leverages Graph Neural Networks (GNNs) \cite{WanZP2022,ZhuYG2023} and self-supervised learning \cite{ZhaYY2023} to handle unseen entities. 
Despite this progress, significant challenges remain. 
As illustrated in Figure \ref{path}, the inherent complexity of KGs—characterized by numerous nodes and diverse relations—often introduces noise during the reasoning process. 
Such irrelevant structural information hinders the effective utilization of key graph characteristics \cite{WanWG2024,jin2024learning,jin2025your}, making robust noise filtering crucial for accurate reasoning.
%The complex KGs contain a large number of nodes and edges with diverse and complex relation types, which may introduce noise during the reasoning process, and hinder the effective exploitation of structural characteristics. Thus, how to effective remove the noise of the irrelevant structural information within the KGs is crucial.

\begin{figure*}[t]
\centering
\includegraphics[width=\linewidth]{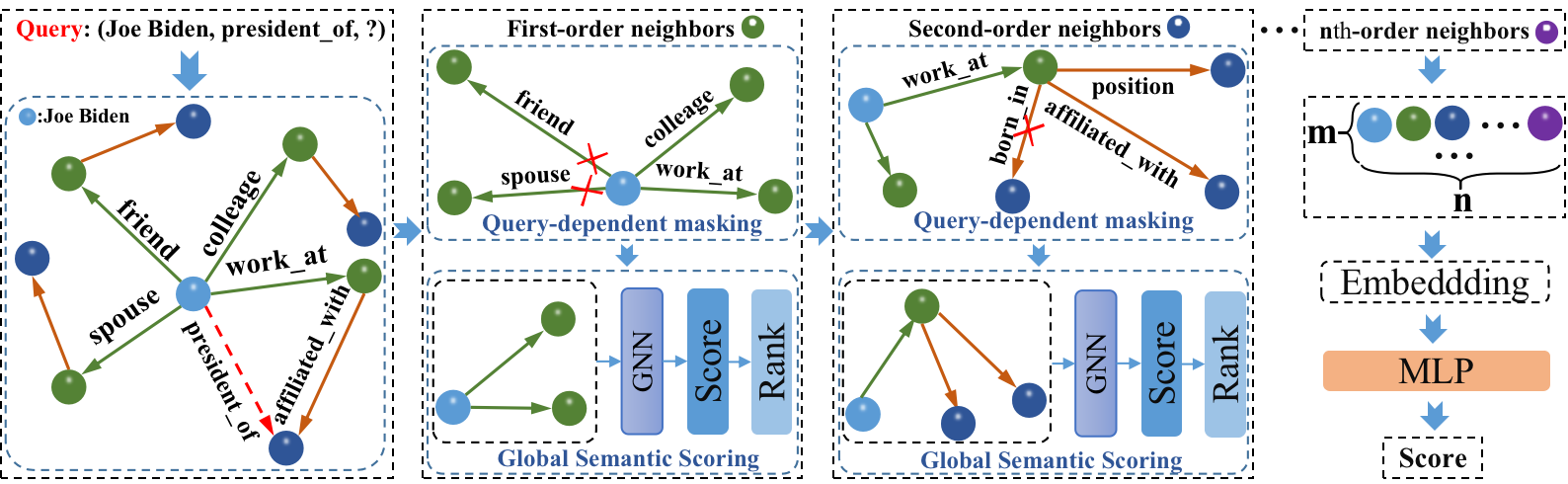}
\caption{Overview of CPSR. Specifically, the query-dependent masking module filters noisy structural information, while the global semantic scoring module captures long-range dependencies across the reasoning path. Finally, an MLP computes candidate scores. Here, \textbf{m} and \textbf{n} denote the number of paths and nodes per path, respectively.}
\label{overview}
\end{figure*}

Furthermore, prior research typically focuses on the current node in a reasoning path, overlooking critical long-range semantic dependencies across the entire sequence. 
This localized perspective restricts the understanding of semantic context, thereby reducing reasoning effectiveness. 
Consequently, it is essential to develop a mechanism that captures the full scope of semantic dependencies by considering the cumulative influence of all nodes along the reasoning path.

%However, existing approaches overlook the significance of these extensive semantic dependencies, resulting in models that are constrained in their ability to perform deep semantic analysis.
To address these challenges, this paper proposes the \textbf{C}umulative \textbf{P}ath-Level \textbf{S}emantic \textbf{R}easoning (CPSR) framework, which simultaneously captures both structural and semantic information of KGs to enhance the inductive KGC task. 
Specifically, CPSR employs a query-dependent masking module to adaptively filter noisy edges while retaining those closely related to targets. 
Furthermore, a global semantic scoring module evaluates both individual node contributions and their collective impact, enabling the accurate identification of critical paths and long-range semantic dependencies.
%that evaluates not only the individual contributions, including influence or centrality, of each node on a path but also their synergistic effects, including enhanced connectivity or reinforced semantic relevance, within the knowledge graph. This precision enables the accurate identification of important semantic paths, thereby effectively capturing long-range semantic dependencies between entities essential for advanced reasoning tasks.

%, which enhances the reasoning process through a refined denoising approach and effectively captures long-range semantic dependencies. Specifically, the framework employs a query-dependent knowledge graph masking (KG mask) mechanism that integrates Bernoulli distribution with single rules(i.e., ordered relation pairs) to retain edges closely related to the targets selectively. Additionally, we have developed a comprehensive path-scoring mechanism that evaluates not only the individual contributions, including influence or centrality, of each node on a path but also their synergistic effects, including enhanced connectivity or reinforced semantic relevance, within the knowledge graph. This precision enables the accurate identification of important semantic paths, thereby effectively capturing long-range semantic dependencies between entities essential for advanced reasoning tasks.
To summarize, the main contributions of this paper are as follows: 
(1) This paper proposes a novel CPSR framework to capture both structural and semantic information of KGs, enhancing the inductive KGC task.
(2) This paper designs a query-dependent masking module, which adaptively masks the noisy structures and retains the relevant structures within KGs in a data-driven manner.
(3) This paper introduces an innovative path global semantic scoring module to capture the long-range semantic dependencies of the reasoning path within KGs.
\section{Preliminary}
\textbf{Inductive Knowledge Graph Completion}. 
A knowledge graph \cite{wang2024ime}, denoted as $\mathcal{G} = (\mathcal{V}, \mathcal{E}, \mathcal{R})$, consists of finite sets of facts (edges) $\mathcal{E}$, entities (nodes) $\mathcal{V}$, and relations  $\mathcal{R}$. Each fact is represented as a triplet $(s, r, o) \in \mathcal{V} \times \mathcal{R} \times \mathcal{V}$, indicating a relation $r$ from the head entity $s$ to the tail entity $o$.
KGC typically predicts the missing information through the known knowledge. Specifically, for a query $(s, r_q, ?)$, the goal is to find the set of answers $\mathcal{V}{(s,r_q,?)}$ such that for all $o \in \mathcal{V}{(s,r_q,?)}$, the triplet $(s, r_q, o)$ holds true. 
Based on KGC, inductive KGC is used to predict missing facts for emerging entities that never appear in KGs. %E′={(s′,q′,o′)|(s′,r′,o′)∉E,s′∈V′oro′∈V′,r′∈RwhereV′⋂V=∅andV′≠∅}\mathcal{E}' = \{(s',q',o')\\|(s',r',o') \notin \mathcal{E}, s' \in \mathcal{V}' \text{or} \; o' \in \mathcal{V}', r' \in \mathcal{R} \; where \; \mathcal{V}' \bigcap \mathcal{V} = \emptyset \; and \;  \mathcal{V}' \neq \emptyset \}.

\textbf{Logical Rules}. Logical rules $\rho$\ \cite{SadAD2019} define the relation between entities $s$ and $o$:
\begin{equation}
\label{logical_rule}
\begin{aligned}
    \rho:\wedge_{i=1}^{l-1}r^{*}(s,o_i) \Rightarrow r_l(s,o),
\end{aligned}
\end{equation}
where the rule body induces ($\Rightarrow$) the head relation $r$.
The rule body is represented by the conjunction ($\wedge$) of a series of body relations $r^* \in \{r_1,...,r_{l-1}\}$. 
The logical rule is referred to as the single rule when $r^*= r_1$. 

% \textbf{Single Rule.} A single rule holds between relation r1r_1 and relation r2r_2 if ∀x,y\forall x,y:
% \vspace{-3pt}
% \begin{equation}
% \label{logical_rule}
% \begin{aligned}
%    r_{1}(x,y) \Rightarrow r_{2}(x,y),
% \end{aligned}
% \end{equation}
% where xx and yy denote the entities respectively.
\section{Method}
Figure \ref{overview} illustrates the CPSR framework, which employs query-dependent masking for noise filtering and global semantic scoring to capture long-range dependencies.
\subsection{Query-Dependent Masking}
The query-dependent masking module filters query-irrelevant noise by extracting rules and normalizing their confidence into probabilities. 
These probabilities then guide Bernoulli sampling \cite{ZhuXY2021} to mask noisy structures and retain critical information. %{\color{red}{This query-dependent masking module allows the model to prioritize highly relevant relations while also considering less relevant ones, thus preserving critical structural information loss and preventing overfitting.}}

Firstly, following logical notations, we denote potential relevance between relation $r$ and relation $r_q$ using a single rule $r \Rightarrow r_q$. 
To quantify relevance, we define rule confidence as:
\begin{equation}
\label{confidence}
\begin{aligned}
    \mathcal{C}(r \Rightarrow r_q) = \frac {\sum_{t \in \mathcal{E}}\mathds{1}(r \in E_{r}(t) \land r_q \in E_{r}(t))}{\sum_{t \in \mathcal{E}} \mathds{1} (r \in E_{r}(t))},\nonumber
\end{aligned}
\end{equation}
where the function $\mathds{1}(x)$ equals 1 when $x$ is true and 0 otherwise, $E_{r}$ denotes the extracted relations from the triplets. 
$\mathcal{C}(r \Rightarrow r_q)$ quantifies the relevance between $r$ and $r_q$; a higher value indicates stronger relevance. 

Next, we calculate the probability based on their confidence. 
The probability is obtained through normalizing confidence into probability:
\begin{equation}
\label{relevance_prob}
\begin{aligned}
    p_{rr_q}^{(l)} = \min(\frac{\mathcal{C}_\text{max}^{(l)}-\mathcal{C}(r \Rightarrow r_q)}{\mathcal{C}_\text{max}^{(l)}-\mathcal{C}_\text{avg}^{(l)}} \cdot p_{e},\ p_{\tau}),
\end{aligned}
\end{equation}
where $p_{rr_q}^{(l)}$ denotes the relevance-based importance of relation $r$; $p_{e}$ is a probability-scaling hyper-parameter; $\mathcal{C}_{\max}^{(l)}$ and $\mathcal{C}_{\text{avg}}^{(l)}$ are the maximum and average confidence scores in $\mathcal{C}^{(l)}$, respectively; and $p_{\tau}$ is a cut-off threshold to prevent the loss of critical relations via truncation.

Finally, we obtain a modified subset $\widetilde{\mathcal{R}}^{(l)}$ from the candidate relations $\hat{\mathcal{R}}^{(l)}$ with certain probabilities at the $l$-th hop:
\begin{equation}
\label{bernoulli_sample}
\begin{aligned}
  \widetilde{\mathcal{R}}^{(l)} = \{r \mid r \in \hat{\mathcal{R}}^{(l)}, Bern(p_{rr_q}^{(l)}) = 1\},
\end{aligned}
\end{equation}
where $\hat{\mathcal{R}}^{(l)}$ denotes the set of $l$-th order neighboring relations of entity $s$, $Bern$ represents a Bernoulli distribution, $\widetilde{\mathcal{R}}^{(l)}$ is then used as the relation set.

After obtaining the subset $\widetilde{\mathcal{R}}^{(l)}$, we can effectively mask the noisy edges, thereby providing a clean KG for the subsequent reasoning process.

\subsection{Global Semantic Scoring}
% Effective path scoring is crucial for capturing long-range semantic dependencies. Accordingly, the proposed global semantic scoring module integrates current and historical node scores to achieve precise path evaluations.

The global semantic scoring module captures long-range dependencies by iteratively integrating current nodes and archiving predecessors, ensuring accurate semantic transmission across complex paths.

During the \(L\)-step reasoning process, every step of the reasoning path with length \(l\) ($0<l\leq L$) treats the node reached at the step as the current node and computes its score:
\vspace{-3pt}
\begin{equation}
\begin{split}
    S_\text{cur}^l=\left\{
                \begin{array}{ll}
                   0, &l=0\\
                  \mathbf{W}^{T}\mathbf{Cur}(l), &1 \leq l \leq L ,\\
                \end{array}
              \right.
\end{split}
\end{equation}
where $\mathbf{W}$ indicates the learnable parameter, $\mathbf{Cur}(l)$ is the embedding of the `current node' with the length $l$.
Furthermore, the reasoning path with length $(l-1)$ $(1\leq l\leq L)$ is treated as historical nodes, and the scores can be computed as follows:
\begin{equation}
\begin{split}
    S(P_{0 \rightarrow(l-1)})=\left\{
                \begin{array}{ll}
                   0, &l=1\\
                  \sum_{i\in\{1,\ l-1\}} S_\text{cur}^{i}, &2 \leq l \leq L ,\\
                \end{array}
              \right.\nonumber
\end{split}
\end{equation}
where $ S_{cur}^{l-1}$ represents the score of the historical node with length $(l-1)$, and $S(P_{0 \rightarrow(l-1)})$ denotes the whole score of all historical nodes.

Unlike methods focusing solely on the current node, our module assesses the entire reasoning path by treating all constituent nodes as integral components, ensuring a more thorough semantic evaluation. 
The detail is shown as follows:
\begin{equation}
\label{scoring_mechanism}
S(P_{0 \rightarrow l}) =
S(P_{0 \rightarrow(l-1)})+ S_\text{cur}^l,
\end{equation}
where $S(P_{0 \rightarrow l})$ indicates the score of the whole reasoning path with length $l$. 

\subsection{Entity Embedding} 
The entity embedding module represents emerging entities by iteratively learning from the \textit{Top}-$k$ reasoning paths selected via global semantic scores.
%Entity embedding encapsulates the rich semantics of entities, which are beneficial for inductive KGC. Consequently, we begin by reviewing path-based approaches and identify their significant computational complexity as a primary limitation. Building on this analysis, we employ a greedy algorithm to effectively address this challenge. Finally, we utilize an iterative learning strategy to acquire precise embeddings for the entities.

Previous research primarily utilizes path-based methods \cite{ZhuZX2021} to represent emerging entities. 
Specifically, they learn a representation $\mathbf{h}_{o}$ for predicting the triplet $(s, r_q, o)$ based on the set of reasoning paths $\mathcal{P}$ from $s$ to $o$:
\begin{equation}
\small
\label{triple_embd}
\begin{aligned}
    \mathbf{h}_{o} &= \sum_{P \in \mathcal{P}}\sum _{(u,r,v)\in P} \mathbf{h}_{(u,r,v)}= \sum_{P \in \mathcal{P}}\sum _{(u,r,v)\in P} \mathbf{W}_{r_q}^{T}[\mathbf{h}_{r_q};\mathbf{h}_{r}],\nonumber
\end{aligned}
\end{equation}
where $[\cdot;\cdot]$ is the concatenation operator, $\mathbf{h}_{(u,r,v)}$ is the representation of triplet $(u, r, v)$  based on the query relation $r_q$, $\mathbf{W}_{r_q}$ is a learnable parameter, $\mathbf{h}_{r_q}$ denotes the representation of the relation $r_q$, and $\mathbf{h}_{r}$ denotes the representation of the relation $r$.

It is worth noting that there are multiple paths to reach the target entity within KGs. %, and exploring a vast path space poses a significant computational challenge for the model. %For reasoning paths of length LL, there are |R|L|\mathcal{R}|^L candidate paths, which markedly amplify the computational complexity. Consequently, we modify Eq ref{iterative_embed} as follows: 
Therefore, we employ a greedy algorithm to select the \textit{Top}-$k$ paths with the highest scores:% and return the end entities associated with these paths.

\begin{equation*}
\label{top_k_nodes}
\begin{aligned}
  \hat{\mathcal{V}}^{(l-1)} = Top\_k (\bigcup_{P \in \mathcal{P}_{0 \rightarrow (l-1)}} S(P_{0 \rightarrow (l-1)})),
\end{aligned}
\end{equation*}
where $\mathcal{P}_{0 \rightarrow (l-1)}$ denotes all the reasoning paths of the length $(l-1)$,
$\hat{\mathcal{V}}^{(l-1)}$ denotes the set of the end entities inferred from $\mathcal{P}_{0 \rightarrow (l-1)}$.
Finally, due to the impacts of the $Top\_k$ strategy adopted by the greedy algorithm on entity embedding, we further revise  as follows:
% \begin{equation}
% \label{modify_scoring_mechanism}
% S(\mathcal{P}_{s \rightarrow t}^{(l)}) =
% \displaystyle \sum\limits_{\substack{(x,r)\in N_{er}^{(1)}(t)\\ x \in \hat{\mathcal{V}}^(l-1),r \in \widetilde{\mathcal{R}}^{(l)}}} \left(S(\mathcal{P}_{s \rightarrow x}^{(l-1)})+Cur(t)\right), \text{if } 1 \leq l \leq L-1
% \end{equation}
\begin{equation}
\small
\mathbf{h}_o^{(l)} = \left\{
\begin{alignedat}{2} 
    & \sum_{(s,r,o) \in \mathcal{E}} \phi, && \hspace{-20pt} l = 1 \\
    & \sum\limits_{x \in \hat{\mathcal{V}}^{(l-1)}} \sum\limits_{(x,r,o) \in \mathcal{E}} \left( \mathbf{h}_{x}^{(l-1)} + \phi \right), 2 \le l \le L,\nonumber
\end{alignedat}
\right.
\label{iterative_embed_for_important_path}
\end{equation}
where $\phi = \mathbf{W}_{r_q}^{T} [\mathbf{h}_{r_q}; \mathbf{h}_{r}]$.

After the above operations, we can obtain the final embedding of the predicated entity $o$.

\subsection{Loss Function} \label{subsec:loss_function}
Following the strategy\ \cite{LacUO2018}, we train the proposed CPSR through the multi-class log-loss function $\mathcal{L}$:
\begin{equation}
\small
\label{multi-class}
    \begin{split}
\mathcal{L} = & \sum_{\mathcal{E}_t} \bigg( -\mathbf{W}_s^T \mathbf{h}_o^{(L)} + \log \Big( \sum_{\forall x \subset \mathcal{V}} \exp \big( \mathbf{W}_s^T \mathbf{h}_x^{(L)} \big) \Big) \bigg),\nonumber
\end{split}
\end{equation}
%\vspace{-3pt}
%\begin{equation}
%\label{scorce}
    %f(s, r_q, o) = \mathbf{w}_{s}^{T}\mathbf{h}_{o}^{(L)},
%\end{equation}
where $\mathbf{W}_{s} \in \mathbb{R}^{d}$ is a weight parameter, $\mathbf{h}_{o}^{(L)}$ represents the embedding of entity $o$ at step $L$, $\mathbf{h}_{x}^{(L)}$ represents the embedding of entity $x$ at step $L$, $\mathcal{E}_{train}$ denotes the set of the positive triplets $(s,r_q,o)$.

\section{Experiments}\label{Sec:Experiments}
\subsection{Experiment setup}
This section describes the datasets, baselines, and experimental configurations used to evaluate CPSR.

\textbf{Datasets.} We adopt eight inductive datasets from \cite{ZhuYG2023}, consisting of four versions of WN18RR and FB15k-237.

\textbf{Baselines.} The proposed CPSR is compared with several classic inductive KGC methods: RuleN\ \cite{MeiFW2018}, NeuralLP\ \cite{YanYC2017}, DRUM \cite{SadAD2019}, GraIL\ \cite{TerDH2020}, NBFNet\ \cite{ZhuZX2021}, RED-GNN \cite{ZhaY2022}, AdaProp\ \cite{ZhaZY2023}, A*Net\ \cite{ZhuYG2023}, and MLSAA \cite{SunJH2024}.

\textbf{Parameter Setting.} CPSR is evaluated by MRR and optimized via Adam, implementation details and hyper-parameters are provided in Appendix \ref{Sec:Parameter}.
%Please refer to the table 2 for more detailed information. The whole experiment is implemented on an NVIDIA RTX 3090 GPU with i9-10900X CPU.

\subsection{Experimental Analysis} 
As shown in Table \ref{tab:results}, our method significantly improves the performance of MRR on most datasets.
This indicates that CPSR can effectively mask the noisy structural information and capture long-range semantic dependencies within KGs through the query-dependent masking and global semantic scoring module, further enhancing inductive reasoning capabilities.

AdaProp\cite{ZhaZY2023} and A*Net\cite{ZhuYG2023} are two important baselines, as they both facilitate the reasoning process by capturing the structural information within KGs. 
However, the proposed CPSR still can obtain the significant improvements on both datasets. 
This phenomenon demonstrates that the query-dependent masking module can effectively mask the noisy structures, thereby enhancing the inductive KGC tasks.

%However, they overlook the impacts of noisy structures and fail to capture long-distance dependencies between entities. Consequently, experimental results demonstrate that the query-dependent masking mechanism and the global semantic scoring mechanism effectively address these issues.
\begin{table}[t]
\centering
\scriptsize
\setlength{\tabcolsep}{2pt}
\renewcommand{\arraystretch}{1.3}

\caption{Performance of inductive KGC on MRR. The best score is in bold, and the second-best score is in underlined.}
\label{tab:results}

\begin{tabularx}{\columnwidth}{l||*4{Y}|*4{Y}}
\hline\hline
\rowcolor[gray]{0.9}
\textbf{Methods} & \multicolumn{4}{c|}{\textbf{WN18RR}} & \multicolumn{4}{c}{\textbf{FB15k237}} \\ \cline{2-9}
\rowcolor[gray]{0.9}
& v1 & v2 & v3 & v4 & v1 & v2 & v3 & v4 \\ \hline\hline
RuleN    & 66.8 & 64.5 & 36.8 & 62.4 & 36.3 & 43.3 & 43.9 & 42.9 \\
NeuralLP & 64.9 & 63.5 & 36.1 & 62.8 & 32.5 & 38.9 & 40.0 & 39.6 \\
DRUM     & 66.6 & 64.6 & 38.0 & 62.7 & 33.3 & 39.5 & 40.2 & 41.0 \\ \hline
GraIL    & 62.7 & 62.5 & 32.3 & 55.3 & 27.9 & 27.6 & 25.1 & 22.7 \\
NBFNet   & 68.4 & 65.2 & 42.5 & 60.4 & 30.7 & 36.9 & 33.1 & 30.5 \\
RED-GNN  & 70.1 & 69.0 & 42.7 & 65.1 & 36.9 & 46.9 & 44.5 & 44.2 \\
MLSAA    & 71.6 & 70.0 & 44.8 & 65.4 & 36.8 & 45.7 & 44.2 & 43.1 \\
AdaProp  & \underline{73.3} & \underline{71.5} & \underline{47.4} & \underline{66.2} & 31.0 & 47.1 & 47.1 & 45.4 \\
A*Net    & 72.7 & 70.4 & 44.1 & 66.1 & 45.7 & \textbf{51.0} & 47.6 & \textbf{46.6} \\ \hline
\rowcolor[HTML]{D7F6FF}
\textbf{CPSR} & \textbf{79.4} & \textbf{79.8} & \textbf{52.0} & \textbf{76.4} & \textbf{48.8} & \underline{49.7} & \textbf{48.8} & \underline{46.3} \\ \hline
\end{tabularx}
\end{table}
 
\subsection{Ablation Experiments}\label{subsec:ablation_studies}
%\textbf{Impact of different modules.} 
%\textbf{Impact of different modules.} 
In order to analyze the impact of each module, we conduct ablation experiments to validate the effectiveness of different modules.
The experimental results are shown in Table \ref{Variants}. 
The experimental results show that our proposed CPSR has a significant improvement on both datasets. 
We conclude that the query-dependent masking module can eliminate the interference of noisy structures, and the global semantic scoring module can fully capture long-range semantic dependencies within the whole reasoning path. %The collaboration of these two mechanisms enhances generalization and robustness.
\begin{table}[t]
\centering
\footnotesize
\renewcommand{\arraystretch}{1.3}
\caption{Ablation studies on both datasets. "CPSR w/o M" and "w/o S" denote the removal of the query-dependent masking and global semantic scoring modules, respectively.}
\label{Variants}
\resizebox{\columnwidth}{!}{
    \begin{tabularx}{1.1\columnwidth}{p{2.2cm}|*3{Y|}*3{Y}} % 稍微加宽一点内部逻辑空间
    \hline\hline
    \rowcolor[gray]{0.9} 
    & \multicolumn{3}{c|}{\textbf{WN18RR(V4)}} & \multicolumn{3}{c}{\textbf{FB15k237(V4)}} \\ \cline{2-7}
    \rowcolor[gray]{0.9} 
    \multirow{-2.3}{*}{ \textbf{Methods} } & MRR & H@1 & H@10 & MRR & H@1 & H@10 \\ \hline\hline
    CPSR w/o M & 73.3 & 62.9 & 94.4 & 45.2 & 35.2 & 63.4 \\ \hline
    CPSR w/o S & 76.2 & 66.7 & 95.4 & 44.6 & 34.5 & 61.8 \\ \hline
    \rowcolor[HTML]{D7F6FF} 
    \textbf{CPSR} & \textbf{76.4} & \textbf{67.2} & \textbf{95.6} & \textbf{46.3} & \textbf{36.1} & \textbf{64.5} \\ \hline
    \end{tabularx}
}
\end{table}

\begin{figure}[h]
\centering
\includegraphics[scale=0.3]{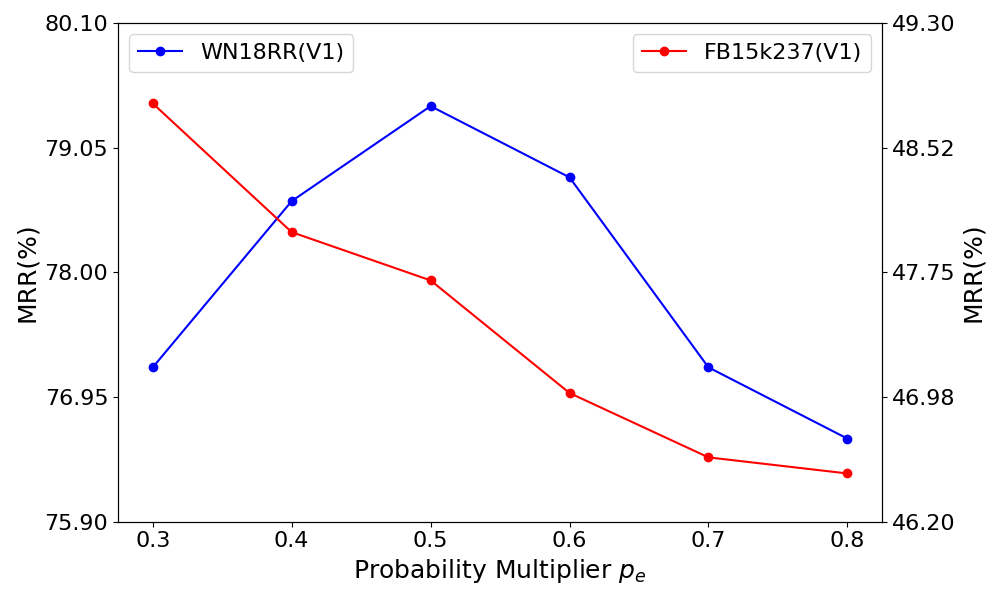}
\vspace{-10pt}
\caption{The performance with different $p_{e}$ values on WN18RR(V1) and FB15k237(V1).}
\label{pm}
\end{figure}
\subsection{Probability Multiplier}
Figure \ref{pm} illustrates the MRR trend for two datasets with different probability multiplier $p_e$.
Overall, the MRR for the WN18RR(V1) dataset first increases and then decreases, showing a clear peak trend. 
In contrast, the MRR for the FB15k237(V1) dataset exhibits a continuous downward trend, with performance gradually weakening as the probability multiplier increases. 
This reflects differences in how datasets handle noise and related information. 
The WN18RR(V1) dataset achieves optimal performance within an intermediate range, while the FB15k237(V1) dataset shows a negative correlation with increasing probability multipliers. 
Therefore, selecting an optimal value of $p_e$ is crucial for maintaining model performance.
\section{Conclusion}
This paper proposes a novel \textbf{C}umulative \textbf{P}ath-level \textbf{S}emantic \textbf{R}easoning (CPSR) framework for inductive KGC tasks. 
Specifically, CPSR first designs a query-dependent masking module to adaptively mask the noisy structural information, ensuring the preservation of critical information.
Subsequently, CPSR develops a global semantic scoring module to capture long-range semantic dependencies in the reasoning path by evaluating the contributions of the current and historical nodes. Experimental results on two widely used datasets show the superiority of CPSR for inductive KGC tasks. 
% \begin{credits}
% % \subsubsection{\ackname} This research is supported by the National Key R\&D Program of China under Grant No.2021ZD0111902; National Natural Science Foundation of China under Grant No.U21B2038, U19B2039, 62172023 and 62206007; R\&D Program of Beijing Municipal Education Commission KZ202210005008, Beijing Natural Science Foundation 4222021, and Engineering Research Center of Intelligent Perception and Autonomous Control, Ministry of Education.
% \end{credits}


\section*{Limitations}
Although CPSR demonstrates robust performance in inductive settings, it has certain limitations. 
First, our path extension relies on a greedy Top-k strategy to balance efficiency and coverage. 
This heuristic may inadvertently prune paths that have low scores in early hops but are essential for the final reasoning, potentially affecting performance on tasks requiring deep, multi-hop logical deductions. Second, the global semantic scoring module requires continuous calculation of semantic relevance relative to the query throughout the path. 
While this effectively alleviates semantic dilution, it introduces additional computational overhead compared to simple local neighbor aggregation methods, particularly when processing entities with extremely high degrees in dense graphs.

\bibliography{sample-base}
\vfill
\break
\appendix
\section*{Appendix}
\section{Parameter Setting} \label{Sec:Parameter}
\begin{table}[!ht] 
\centering
\footnotesize
\renewcommand{\arraystretch}{1.2}

\caption{Hyper-parameter configurations of CPSR on both datasets. $\mathcal{H}$ denotes the hyper-parameter, and “BS” is the batch size.}
\label{inductive_setting}
\resizebox{\columnwidth}{!}{
    \begin{tabularx}{1.1\columnwidth}{c||*4{Y|}*4{Y}}
    \hline\hline
    \rowcolor[gray]{0.9} 
     & \multicolumn{4}{c|}{\textbf{WN18RR}} & \multicolumn{4}{c}{\textbf{FB15k237}} \\ \cline{2-9}
    \rowcolor[gray]{0.9} 
    \multirow{-2}{*}{$\mathcal{H}$} & V1 & V2 & V3 & V4 & V1 & V2 & V3 & V4 \\ \hline\hline

    $L$         & 3    & 3    & 7    & 3    & 7    & 3    & 7    & 5    \\ 
    $K$         & 150  & 50   & 100  & 300  & 300  & 250  & 300  & 300  \\ 
    $p_e$       & 0.5  & 0.3  & 0.3  & 0.6  & 0.3  & 0.7  & 0.3  & 0.4  \\ 
    $p_{\tau}$  & 0.5  & 0.5  & 0.5  & 0.5  & 0.5  & 0.5  & 0.5  & 0.5  \\ 
    BS          & 100  & 50   & 100  & 10   & 20   & 10   & 20   & 20   \\ \hline
    \end{tabularx}
}
\end{table}

The proposed CPSR chooses \textit{Mean Reciprocal Rank} (MRR)\ \cite{WanWG2023} as evaluation metric.
Additionally, we adopt Adam\cite{KinB2015} as the optimizer.
Furthermore, we tune the length of reasoning path $L$, the number of selected paths $K$ in entity embeddings,  probability multiplier $p_e$, cut-off probability $p_{\tau}$ in query-dependent masking module, and list the detailed information of the hyper-parameter in Table \ref{inductive_setting}.
\end{document}